# From 100,000+ images to winning the first brain MRI foundation model challenges: Sharing lessons and models

Pedro M. Gordaliza[1,2,†], Jaume Banus[2†], Benoît Gérin[3], Maxence Wynen[3,4], Nataliia Molchanova[2,5], and Jonas Richiardi[2,1‡], Meritxell Bach Cuadra[1,2‡]

[1]CIBM Center for Biomedical Imaging, Switzerland, [2]Department of Radiology, Lausanne University Hospital (CHUV) and University of Lausanne (UNIL), [3]ICTEAM, Universite Catholique de Louvain, Louvain-la-Neuve, [4]Neuroinflammation Imaging Lab (NIL), Universite Catholique de Louvain, Brussels, [5]University of Applied Sciences Western Switzerland (HES-SO). [†,‡]These authors contributed equally to this work

**Abstract**

Developing Foundation Models for medical image analysis is essential to overcome the unique challenges of radiological tasks. The first challenges of this kind for 3D brain MRI, SSL3D and FOMO25, were held at MICCAI 2025. Our solution ranked first in tracks of both contests. It relies on a U-Net CNN architecture combined with strategies leveraging anatomical priors and neuroimaging domain knowledge. Notably, our models trained 1-2 orders of magnitude faster and were 10× smaller than competing transformer-based approaches. [Models are available here](#).

**Article**

Foundation models (FM) have revolutionized artificial intelligence[1], first in natural language processing[2] (e.g., GPT, BERT) and subsequently in computer vision[3] (e.g., JEPA, DINO). These models, pre-trained on massive datasets using self-supervised learning (SSL), enable fine-tuning for diverse downstream tasks with minimal labeled data, marking a paradigm shift from training task-specific models from scratch. Medical imaging stands to benefit enormously from this approach[4]. Radiology faces persistent challenges: institutional data sparsity, protocol variability and expensive expert annotations result in datasets insufficient to characterize biological heterogeneity. Brain Magnetic Resonance Imaging (MRI) exemplifies both the promise and difficulty of FM in medical imaging. The brain's anatomical complexity and wide spectrum of neurological, oncological, and psychiatric pathologies, offer an ideal testbed for generalizable models. However, the field faces unique obstacles: high-dimensional 3D volumetric data, complementary MRI contrasts, vendor-specific acquisition protocols, and population heterogeneity[5]. Until recently, no standardized benchmarks existed to rigorously evaluate FM capacity to overcome these barriers.

This gap motivated two MICCAI 2025 challenges: the SSL for 3D Medical Imaging Challenge (SSL3D)[6] and the Foundation Model Challenge for Brain MRI (FOMO25)[7]. These competitions represented the first rigorous evaluation of SSL FM for neuroimaging. SSL3D assembled an unprecedented pre-training dataset of 34,191 subjects with multiple contrasts and timepoints, totaling 114,570 3D volumes spanning over 800 heterogeneous datasets, with available clinical metadata (sex, age, health status) for some subjects. The challenge evaluated few-shot generalization across four segmentation and three classification tasks, with organizers fine-tuning submitted pre-trained models. FOMO25 prepared 11,187 subjects totaling 60,529 3D volumes and tested models across segmentation, classification, and regression, with participants submitting directly fine-tuned models from a common pre-trained FM. Together, these challenges created the first standardized arena for comparing FM strategies.

Our team achieved top performance in both competitions by exploiting a key principle: neuroimaging data contains intrinsic structural priors that enable more effective representation learning than generic SSL approaches. We employed CNN-based U-Net architectures trained with masked autoencoders (MAE), guided by a common learning principle implemented in two challenge-specific variants[6–9]. Our core strategy consisted in disentangling subject-invariant anatomical representations from contrast-specific pathological features. Rather than learning monolithic embeddings, we explicitly induced the models to capture: (1) subject-specific anatomical features, consistent across contrasts and timepoints, and (2) contrast-dependent representations encoding pathology visible only in certain contrasts. This inductive bias aims to prevent spurious correlations and shortcut learning by anchoring learning to domain knowledge.

For SSL3D, we partitioned the learned representations into two components: one constrained to match T1-weighted anatomical segmentations across all images of a subject, enforcing consistency across contrasts and timepoints, and another optimized to discriminate subject health status using contrast-specific pathology labels. Both components were combined in the decoder for MAE reconstruction. For FOMO25, we similarly structured the latent space into subject-invariant anatomical features and contrast-specific information. During pre-training, alongside MAE reconstruction, we implemented a cross-contrast reconstruction objective by swapping representations between contrasts of the same subject while maintaining a shared anatomical component. This encourages the model to disentangle anatomy from acquisition-specific characteristics, preserving contrast-specific information for downstream tasks.

In both challenges, our models achieved top average performance across downstream tasks. While no single method dominated every task, a distinct pattern emerged: CNN-based architectures systematically outperformed transformer-based submissions (e.g., 2.5% higher average Dice for segmentation and 8% higher accuracy for classification in SSL3D). Moreover, the efficiency advantage was substantial. In FOMO25, our CNN model required ~36 GPU-hours (<80GB vRAM) for pre-training compared to ~100-1000 hours for comparable transformer approaches, with far fewer parameters (20M vs 300M for ViT-L DINOv2 3D). Notably, newer architectures like DINOv3 (7B parameters) showed limited transfer performance on medical imaging tasks.

These results raise a critical question: why do transformers, which have revolutionized natural image analysis[3], consistently underperform CNNs in medical imaging? In both challenges, not a single transformer-based submission matched top CNN methods. This finding aligns with recent systematic reviews[10] showing that U-Net and its variants continue to achieve state-of-the-art results on most 3D medical image segmentation benchmarks. Though likely confounded, these factors point to fundamental mismatches between transformer architectures and volumetric medical imaging. First, transformers require substantially larger training corpora to learn effective attention patterns, these results suggest that even datasets exceeding 100,000 volumes are insufficient for capturing long-range dependencies in this domain[1,3,4]. Second, 3D tokenization creates severe computational bottlenecks. Treating volumetric images as sequences of 3D patches generates massive token counts; computing pairwise attention yields quadratic complexity that fundamentally limits the spatial resolution and context window transformers can process. Third, fine-tuning transformers for medical imaging tasks remains fragile. Classification requires careful feature aggregation, segmentation needs adapted upsampling mechanisms, and regression demands precise feature extraction—task-specific modifications that undermine the FM universality promise. Combined with computational overhead, these limitations make current vision transformer approaches impractical for 3D medical imaging under resource-constrained settings.

Whether transformers will eventually match CNN performance with larger datasets, more efficient attention mechanisms, or hybrid architectures remains uncertain[1,4]. However, one lesson emerges clearly: FM success depend less on architectural novelty or parameter scale than on principled exploitation of neuroimaging rich structure. Models that explicitly leverage domain structure—longitudinal trajectories, complementary contrasts, and anatomical priors—achieve superior performance and efficiency compared to purely data-driven approaches[5]. This advantage may persist even as datasets scale, or it may diminish as Richard Sutton's "bitter lesson" suggests; nevertheless, leveraging medical imaging structure accelerates progress now, rather than waiting for a breakthrough. Equally critical is developing standardized evaluation frameworks beyond these initial challenges—benchmarks assessing not just accuracy but robustness to domain shift, uncertainty quantification, and performance on rare pathologies. Clinical translation demands models with acceptable computational costs, reliable uncertainty estimates, and robust cross-site performance capabilities. Our winning approaches demonstrate these capabilities are achievable today. The path forward requires neither architectural dogmatism nor uncritical hype, but rigorous evaluation, honest acknowledgment of what remains uncertain, and principled exploitation of the unique properties that distinguish medical imaging from natural images. All pre-trained models, weights, and code are openly available at [github.com/jbanusco/BrainFM4Challenges](github.com/jbanusco/BrainFM4Challenges), enabling the community to build upon this foundation.

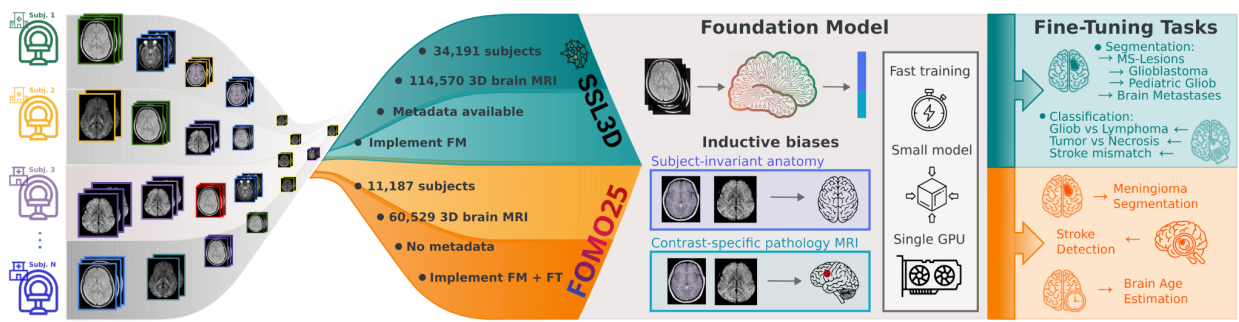

**Figure:** Overview of the MICCAI 2025 SSL3D and FOMO25 challenges and our top-performing Foundation Model strategy. The diagram summarizes the large-scale heterogeneous datasets used for pre-training (left) and the specific constraints of each competition[6–9]. Our methodology (center) leverages inductive biases to disentangle subject-invariant anatomy from contrast-specific pathology, allowing for the training of lightweight, efficient CNN models on a single GPU. These models demonstrated superior generalization across diverse downstream tasks (right)—ranging from glioblastoma segmentation to brain age estimation—achieving 1st place in tracks of both leaderboards.